\documentclass{article}
\usepackage{spconf,amsmath,graphicx}
\usepackage{tabularx,url}

\title{Sketch2Manga: Shaded Manga Screening from Sketch with Diffusion Models}
\name{Jian Lin$^1$, Xueting Liu$^{*1}$, Chengze Li$^1$, Minshan Xie$^2$, Tien-Tsin Wong$^2$\thanks{$^*$Corresponding author (e-mail: tliu@cihe.edu.hk). This work was supported by the Research Grants Council of the Hong Kong SAR, China (Project No. UGC/FDS11/E01/21).}}
\address{$^1$School of Computing and Information Sciences, Saint Francis University\\
$^2$Department of Computer Science and Engineering, The Chinese University of Hong Kong}
\begin{document}
%
\maketitle
\begin{abstract}
While manga is a popular entertainment form, creating manga is tedious, especially adding screentones to the created sketch, namely \textit{manga screening}. Unfortunately, there is no existing method that tailors for automatic manga screening, probably due to the difficulty of generating high-quality shaded high-frequency screentones. The classic manga screening approaches generally require user input to provide screentone exemplars or a reference manga image. The recent deep learning models enables the automatic generation by learning from a large-scale dataset. However, the state-of-the-art models still fail to generate high-quality shaded screentones due to the lack of a tailored model and high-quality manga training data. In this paper, we propose a novel sketch-to-manga framework that first generates a color illustration from the sketch and then generates a screentoned manga based on the intensity guidance. Our method significantly outperforms existing methods in generating high-quality manga with shaded high-frequency screentones.
\end{abstract}
\begin{keywords}
manga generation, manga screening, sketch-to-manga
\end{keywords}
\section{Introduction} 
Manga is a popular form of entertainment that presents stories through black-and-white illustrations. Due to the lack of color, screentones (black-and-white patterns, such as dots, stripes, etc.) are used to present shading and textures. To produce manga pages, the artists usually first create a sketch to depict the contour of the objects and then apply screentones to the sketch at proper places with proper shading. While drawing the contour of the objects comes from the creative thoughts of the artists, the screening process is tedious and repetitive, with a goal of achieving natural yet visually striking imitations of color shades and conveying semantic meanings. An automatic manga screening approach is highly desirable to reduce labor cost, yet unfortunately, this task is extremely challenging because of the large information gap between the sketch domain and the manga domain. The selection and application of screentones pose significant challenges due to the bitonal nature of manga, requiring careful aesthetic and semantic considerations. Additionally, it is not easy to generate black-and-white screentones of high-frequency regular patterns, even with recent deep learning models.


In this work, we propose a novel diffusion-based framework to facilitate the automatic screening process by establishing color illustrations as an intermediary between sketches and manga. Utilizing color illustrations as a proxy simplifies the complex screening task by breaking it down into two more manageable sub-tasks. In the first step, we generate a color illustration from the sketch that carries shading information with a text-to-image diffusion model with line conditioning. In the second step, we generate a manga image with high-frequency screentones from the shaded color illustration. Considering the ultimate goal of mimicking color shading with screentones, the transition from color illustrations to manga is both intuitive and meaningful.
To our knowledge, there is currently no fully automatic approach designed specifically for manga screening from sketches. Existing methods primarily concentrate on photo-to-manga conversion techniques~\cite{qu2008richness,li2011content,wu2014mangawall,xie2020manga,zhang2021generating}, which still present significant domain gaps, or on reference-based screening methods~\cite{wu2023shading,li2023reference} that require additional reference inputs.
While recent advancements in diffusion-based image synthesis have demonstrated impressive capabilities in terms of generation quality and controllability through text~\cite{latentdiffusion} or other conditional inputs~\cite{zhang2023adding}, diffusion models inherently struggle to produce high-quality screentones. This limitation is mainly attributed to the lack of high-fidelity manga data available for training latent diffusion models including both the variational autoencoder (VAE) and the U-Net model, resulting in a propensity to generate visuals with low-frequency characteristics.

\begin{figure*}[!t]
    \centering
    \includegraphics[width=\linewidth]{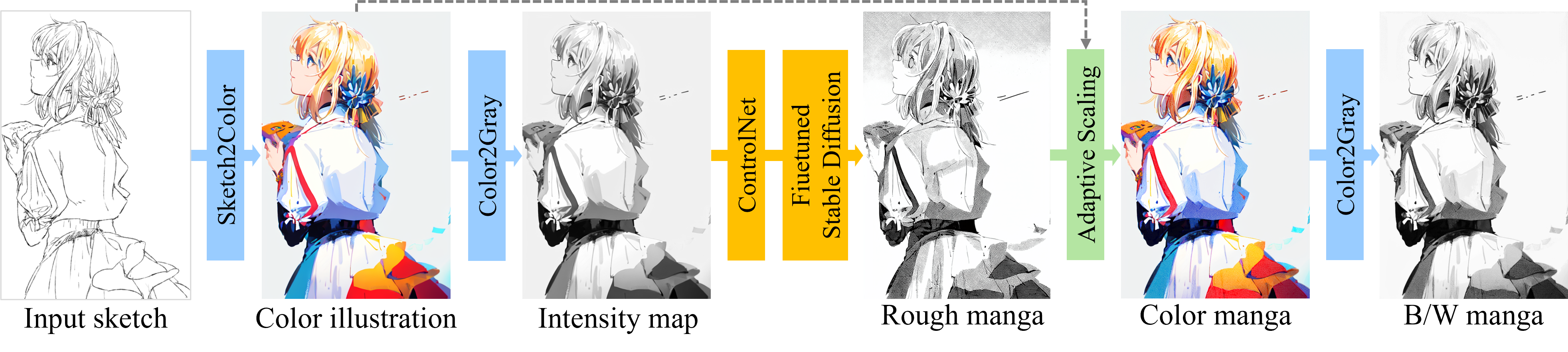}
    \vspace{-0.35in}
    \caption{System overview. Given an input sketch, our system first generates a color illustration from the sketch and then generates a rough manga image using a diffusion model conditioned on the intensity map of the color illustration. An adaptive scaling process is further applied to generate the final manga image.}
    \label{fig:flowchart}
\end{figure*}

In light of these challenges during manga generation, our framework first improves the Stable Diffusion model by finetuning its components with a newly curated high-quality manga dataset. Specifically, we revisit and reformulate the training objectives of its VAE model to better adapt to the intricacies of high-frequency screentones. For the denoise diffusion U-Net model, we utilize the intensity of color illustrations as the additional controlling factor besides text prompts to ensure that the screentones generated for the manga align with the shading of the color illustrations.
Despite the advancements achieved through finetuning, the manga images occasionally exhibit minor structural distortions and unnatural shadings, a byproduct of the denoise diffusion process. To rectify these issues, we have developed an adaptive scaling method that effectively incorporates the generated screentones into the color illustrations, culminating in a manga output that is both structurally sound and aesthetically pleasing.  


We validate the effectiveness of our method by comparing with both sketch-to-manga methods and illustration-to-manga methods. The results show that our method significantly outperforms existing methods. Our contributions are summarized as follows.
\begin{itemize}
\vspace{-0.1in}
    \item We propose a novel sketch to manga framework that generates high-quality manga with shaded high-frequency screentones via a two-step approach with a color illustration serving as the intermediary.
\vspace{-0.1in}
    \item We finetune a diffusion model with a tailored loss function and intensity conditioning to generate high-quality screentones.
\vspace{-0.1in}
    \item We propose an adaptive scaling method to integrate screentones into a color illustration to obtain the manga.
\vspace{-0.1in}
\end{itemize}

\section{Related Work}
\textbf{Sketch-to-manga} While various sketch colorization methods (e.g.,~\cite{zhang2018two,zou2019language,zhang2021user}) have been proposed, very few attempts have been made in adding screentones to sketches. To enable automatic manga screening, \cite{tsubota2019synthesis} proposed to first predict a screentone category label for each pixel via a neural network and then apply the corresponding screentone to each screentone category area. Due to the unstable performance of the classification network, the generated screentone category label map is error-prone and the generated screentones are unnatural and unshaded. Some reference-based manga screening methods have been proposed to transfer the screentones from a manga image to a line drawing \cite{wu2023shading,li2023reference}, but these methods rely on additional information to apply the screentones and synthesize the manga image, which are by nature not automatic. In this paper, we propose a fully automatic method without relying on any additional information.

\vspace{0.1in}
\noindent\textbf{Illustration-to-manga} Despite limited research efforts in sketch-to-manga, many more efforts have been made in color-to-manga. The first photo-to-manga approach~\cite{qu2008richness} proposed to analyze the correspondence between colors and screentones, and then convert color to screentones based on manually selected screentone exemplars. To avoid manual screentone selection, some attempts~\cite{li2011content,wu2014mangawall} have been made to directly generate screentones. Without reference screentone exemplars, the generated screentones are usually unnatural and of low quality. With the development of deep learning networks, a screentone encoder, namely ScreenVAE~\cite{xie2020manga}, has been proposed to allow the inter-transfer of color illustrations and manga. However, this screentone encoder still fails to reconstruct high-quality screentones, especially in rich-color regions. Later, another deep learning approach~\cite{zhang2021generating} has been proposed to estimate the screentone segmentation map and then generate high-frequency screentones based on the segmentation map and screentones exemplars. While this method achieves state-of-the-art performance, the generated manga still cannot preserve the shading of the original image and shows consistently low-contrast. In comparison, we propose a method that can generate high-quality manga with well-shaded high-frequency screentones.

\begin{figure*}[!t]
    \centering
    \begin{minipage}[b]{\linewidth}
        \includegraphics[width=0.165\linewidth]{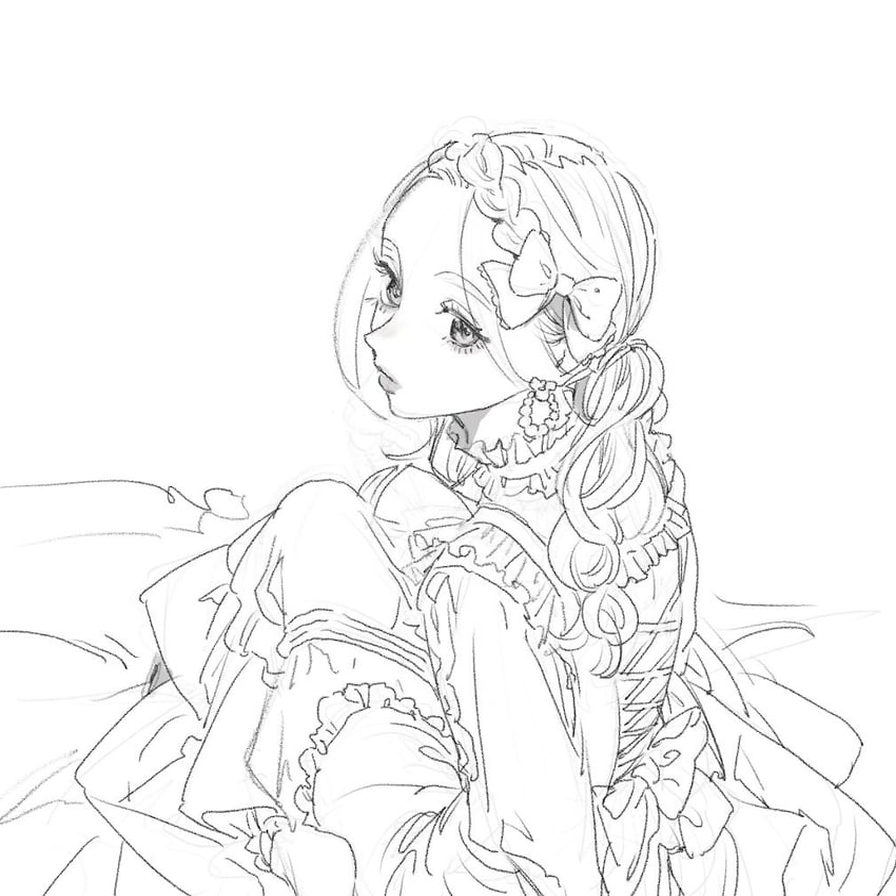}\hfil
        \includegraphics[width=0.165\linewidth]{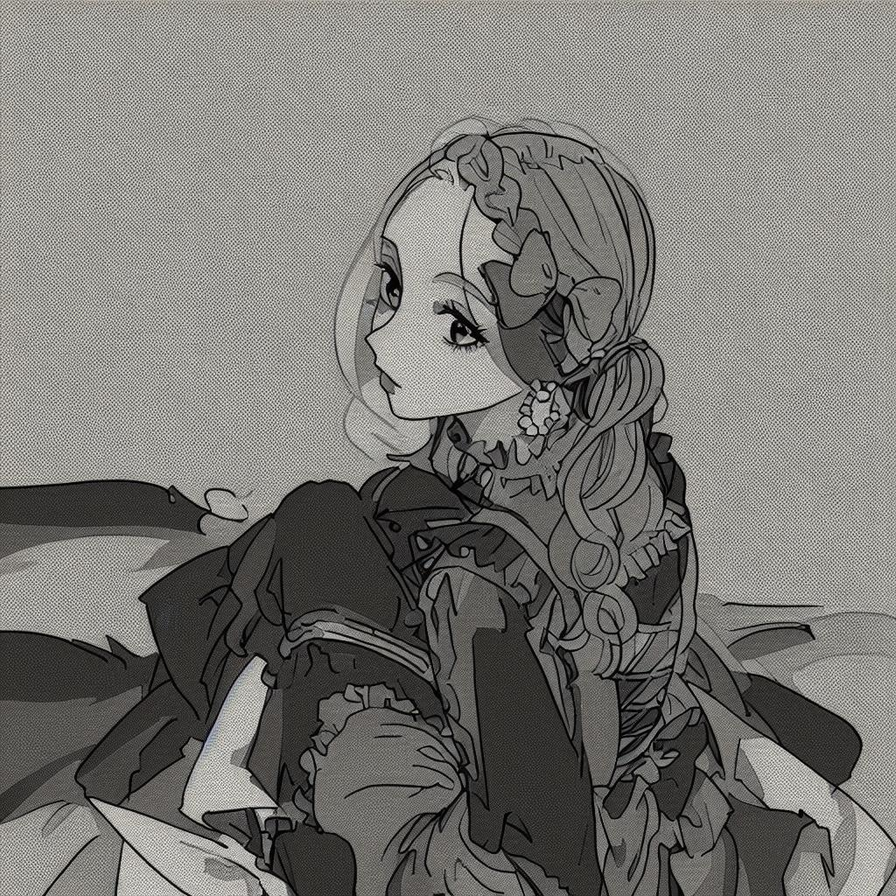}\hfil
        \includegraphics[width=0.165\linewidth]{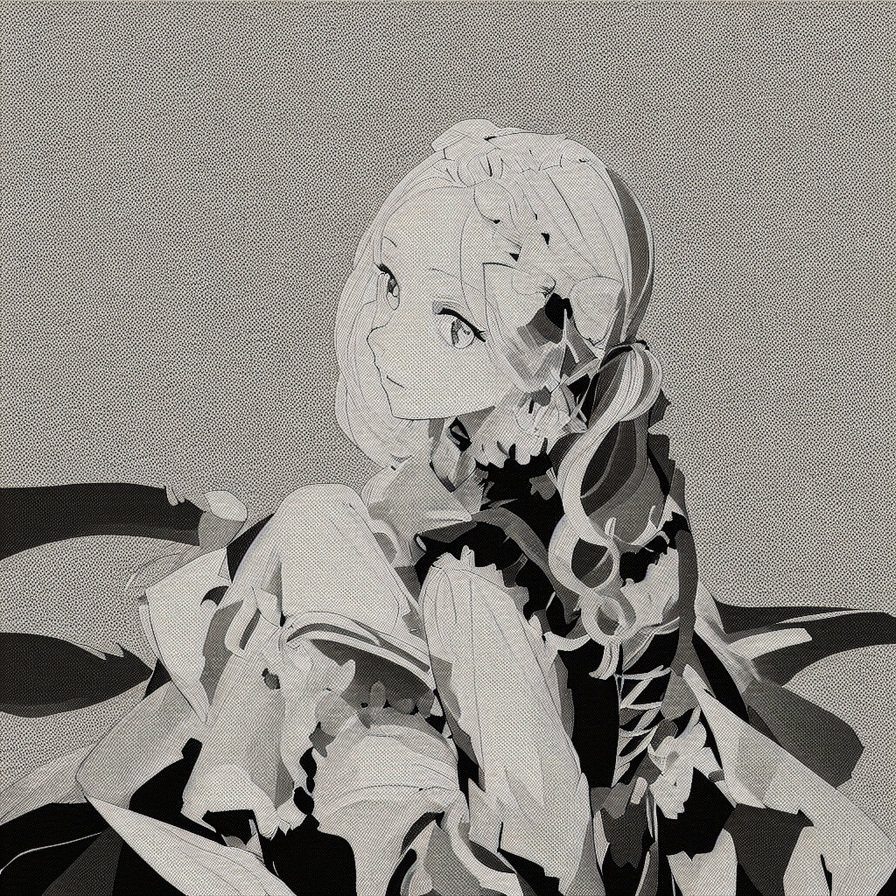}\hfil
        \includegraphics[width=0.165\linewidth]{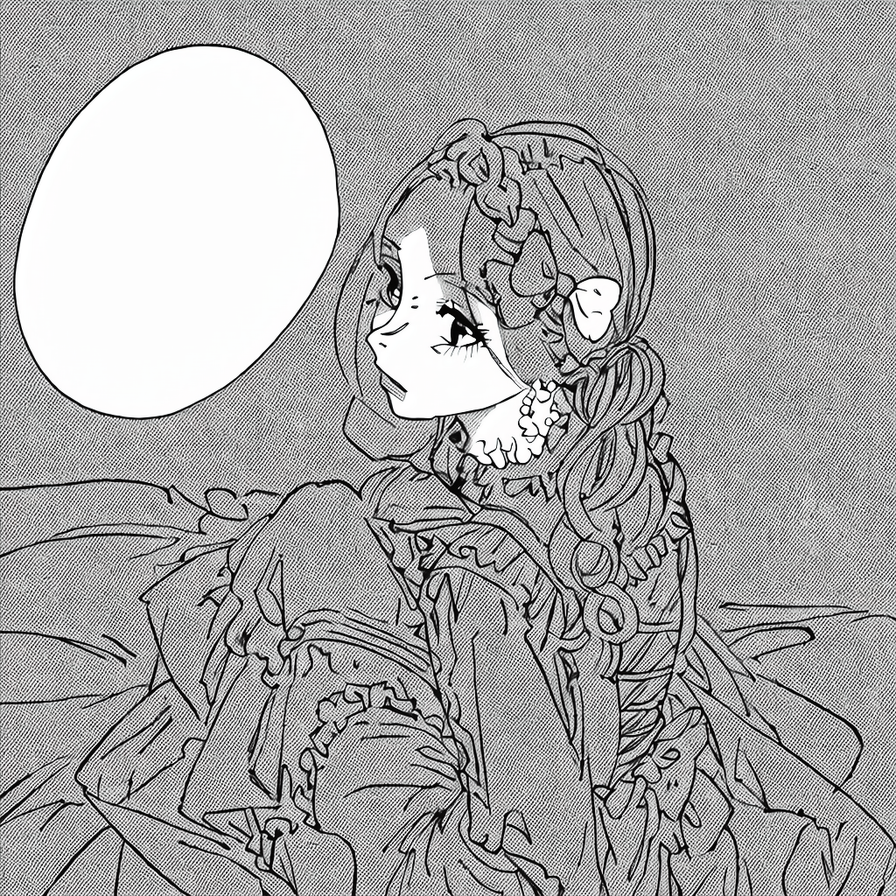}\hfil
        \includegraphics[width=0.165\linewidth]{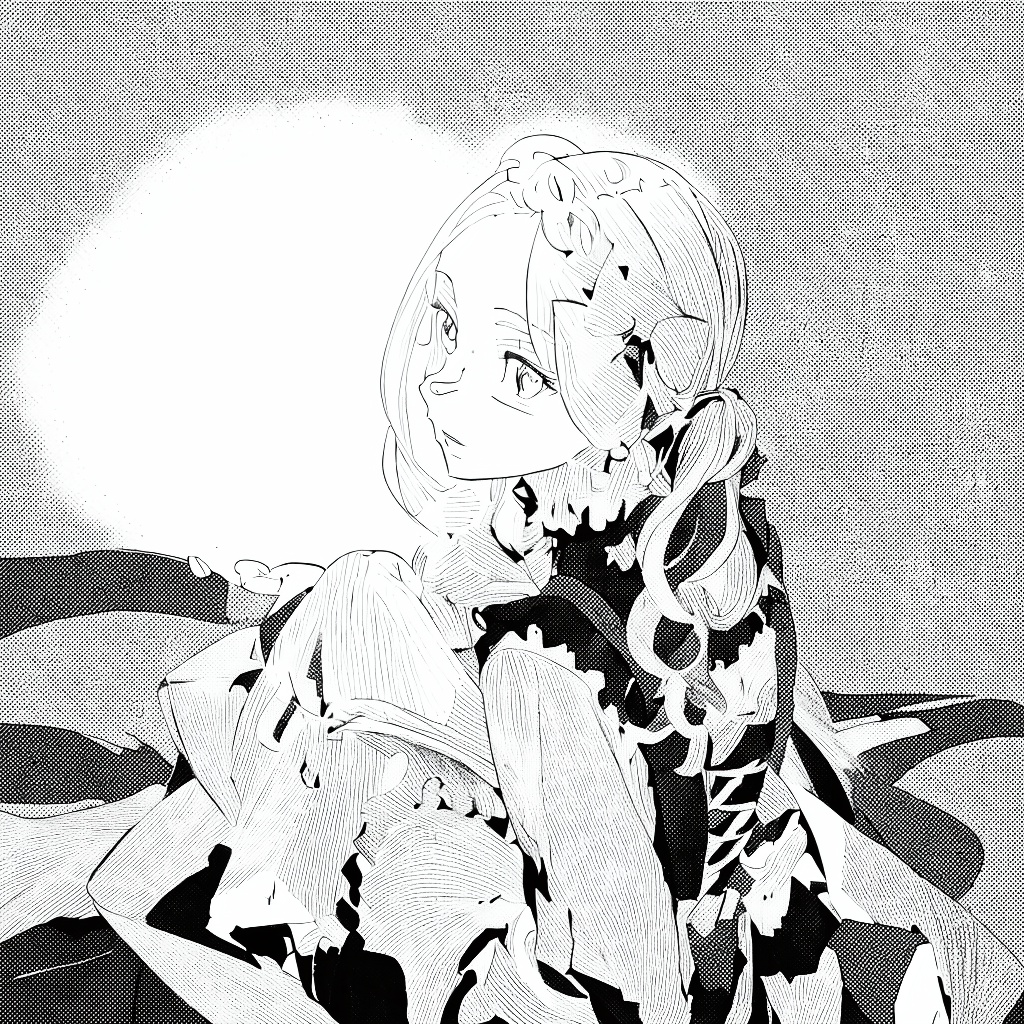}\hfil
        \includegraphics[width=0.165\linewidth]{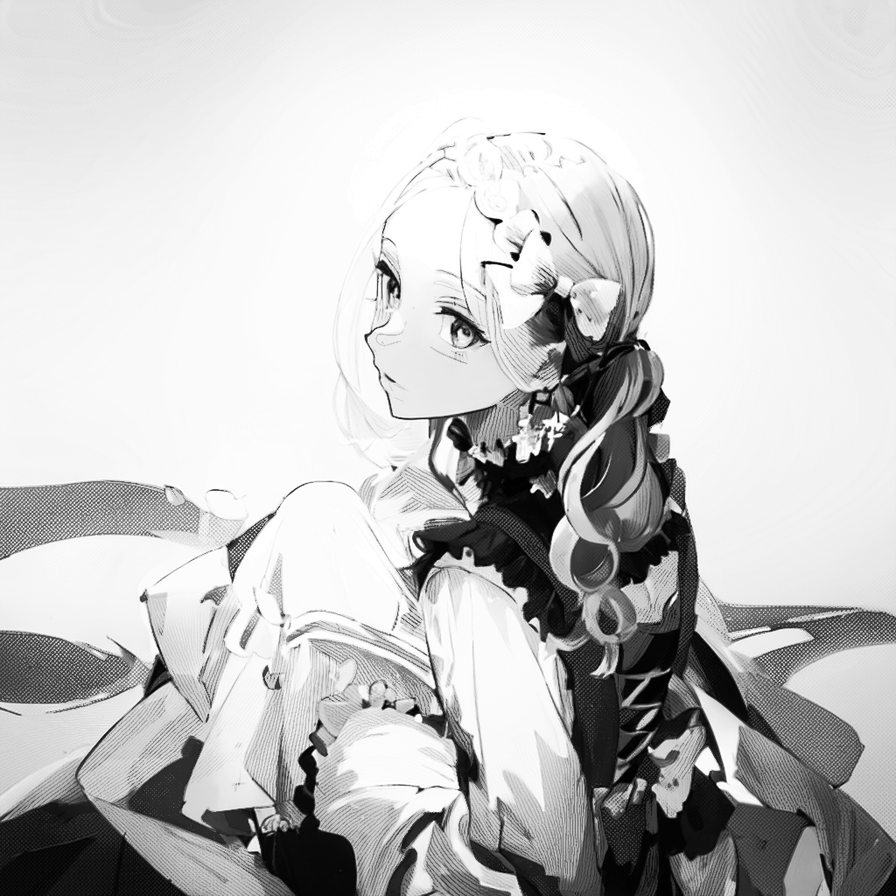}
    \end{minipage}
    \begin{tabularx}{\textwidth}{*{6}{>{\centering\arraybackslash}X}}
            (a) Input sketch & 
            (b) Original cond. on sketch & 
            (c) Original cond. on intensity &
            (d) Finetuned cond. on sketch &
            (e) Finetuned cond. on intensity &
            (g) Output manga based on (e)
    \end{tabularx}
    \vspace{-0.15in}
    \caption{Ablation on latent diffusion. The original diffusion model fails to generate high-frequency screentones without finetuning. After finetuning, conditioning on sketch still leads to flat shading, while conditioning on intensity generates shaded high-frequency screentones.}
    \label{fig:latentdiffusion}
\end{figure*}

\vspace{0.1in}
\noindent\textbf{Diffusion Models and ControlNet} Image generation has attracted numerous attention during the development of deep learning models. Among the existing models, the diffusion models have been proved to be successful in image generation, which are trained to iteratively denoise noise-corrupted samples drawn from Gaussian distribution, resulting in high-quality data samples. The denoising process is mainly performed in latent space \cite{latentdiffusion} for computational efficiency, and the diffusion modules are often incorporated with cross-attention modules that process text embeddings to allow text-to-image generation.
ControlNet~\cite{zhang2023adding} conditions the latent diffusion model on various conditioning signals, such as sketch, to ensure the generation process to closely follow the conditioning signal.
However, the state-of-the-art stable diffusion model with variational autoencoder still generates distorted and blurry screentones when used to generate manga images from the latent space due to significant information loss~\cite{zhu2023designing}. Furthermore, directly generating a manga image using the stable diffusion model conditioned on sketch leads to bland results due to the lack of information in the sketch guidance. In this paper, we propose to finetune the stable diffusion model with high-quality manga data. We also propose a two-step approach to first estimate the color illustration of the sketch and then generate a high-quality manga image with rich screentones via the finetuned stable diffusion model conditioned on the intensity map.

\section{Methodology}
Figure~\ref{fig:flowchart} provides a schematic representation of the architecture of our framework.

\subsection{Sketch to color illustration}

We first propose to generate a color illustration from the input sketch with off-the-shelf diffusion models finetuned on color illustrations from the community, denoted as \emph{Sketch2Color} in Fig.~\ref{fig:flowchart}. Specifically, we use MeinaPastel\footnote{https://civitai.com/models/11866/meinapastel}, a diffusion model finetuned on colored illustration, as our colorization model, and control\_v11p\_sd15\_lineart\footnote{https://github.com/lllyasviel/ControlNet-v1-1-nightly} to condition the denoising process on line-art sketches. We choose the same set of text prompts\footnote{Positive: \emph{masterpiece, best quality, solo, illustration, simple background}; Negative: \emph{nsfw, nude, lowres, bad anatomy, bad hands, worst quality, low quality, normal quality, jpeg, jpeg artifacts}.} for all generations. The generated color illustration is richer in information than the original sketch, encompassing elements such as color and shading. Subsequently, we obtain the intensity values derived from the grayscale version of the color illustration to guide the manga generation with another diffusion model, denoted as \emph{Color2Gray} in Fig.~\ref{fig:flowchart}.

\subsection{Intensity-guided manga generation}

To create manga images with screentones from a sketch, we can use a pre-trained ControlNet~\cite{zhang2023adding} with line conditioning to help guide the generation process with a stable diffusion model~\cite{latentdiffusion}. However, there's a challenge -- many manga images found online are of low quality, often due to being scanned or being compressed for sharing. This means that high-quality screentone images are rare during model pretraining, which leads to lower quality in generating manga. Moreover, we discover that conditioning ControlNet on the grayscale intensity map derived from the colored illustration obtained in the previous step results in a broader range of screentone variations that align more closely with the concept of shading, as opposed to conditioning solely on the sketch.

To obtain a large number of high-resolution manga training samples, we scraped 186k high-resolution manga images from the Internet and further restored the high-resolution versions of the Manga109 dataset~\cite{manga109} using a manga restoration method~\cite{xie2021exploiting} to obtain 103k more high-quality manga images. Altogether 289k high-resolution manga images are collected. Since the prepared manga images are usually full pages containing multiple mangas, the panels are cropped using Grounded DINO (with query \emph{panel}) as training data. The prompts used for text-to-image generation are generated by a SwinV2 tagger\footnote{https://github.com/SmilingWolf/SW-CV-ModelZoo} pretrained on Danbooru~\cite{danbooru2021} dataset.

\begin{figure}[!t]
    \centering
    \begin{minipage}[b]{\linewidth}
        \includegraphics[width=0.48\linewidth]{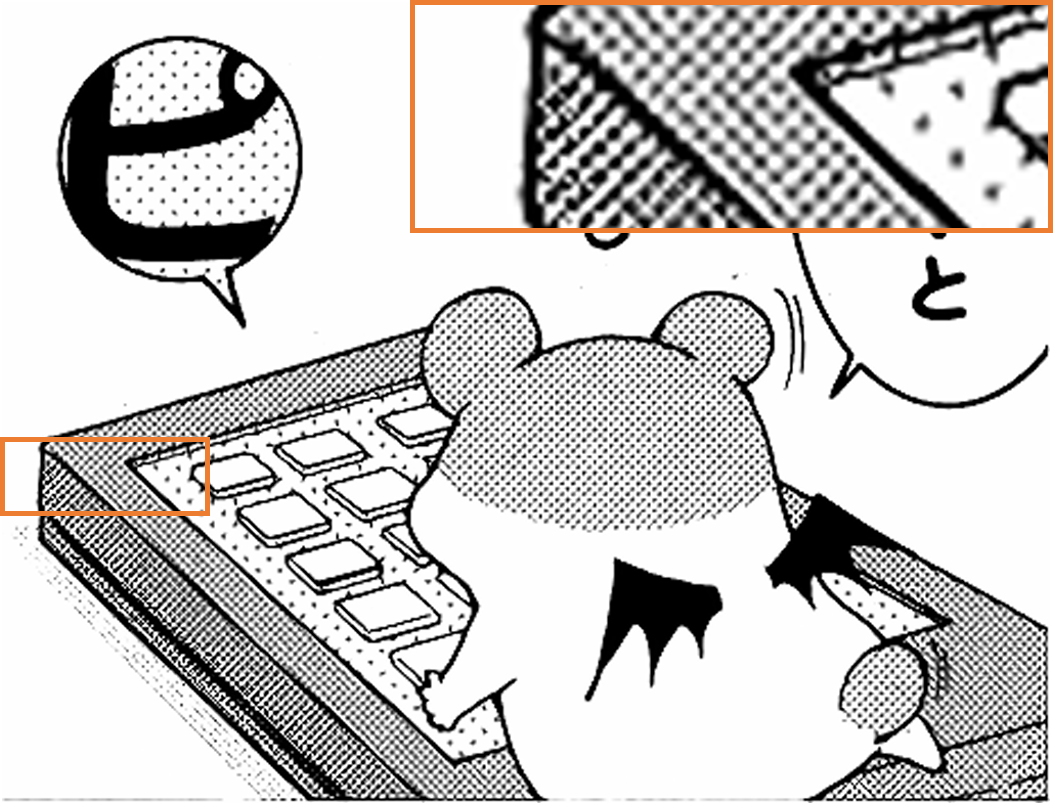}\hfil
        \includegraphics[width=0.48\linewidth]{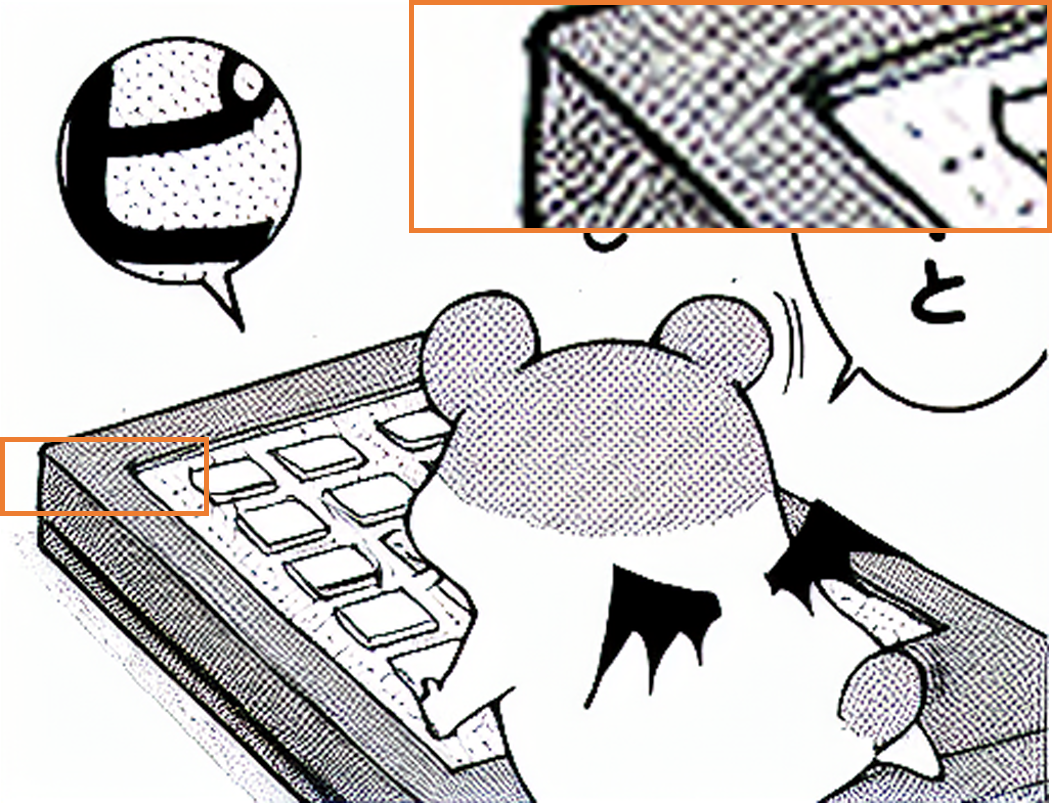}
    \end{minipage}
    \begin{tabularx}{0.48\textwidth}{*{2}{>{\centering\arraybackslash}X}}
            (a) Input & 
            (b) Original
    \end{tabularx}
    \begin{minipage}[b]{\linewidth}
        \includegraphics[width=0.48\linewidth]{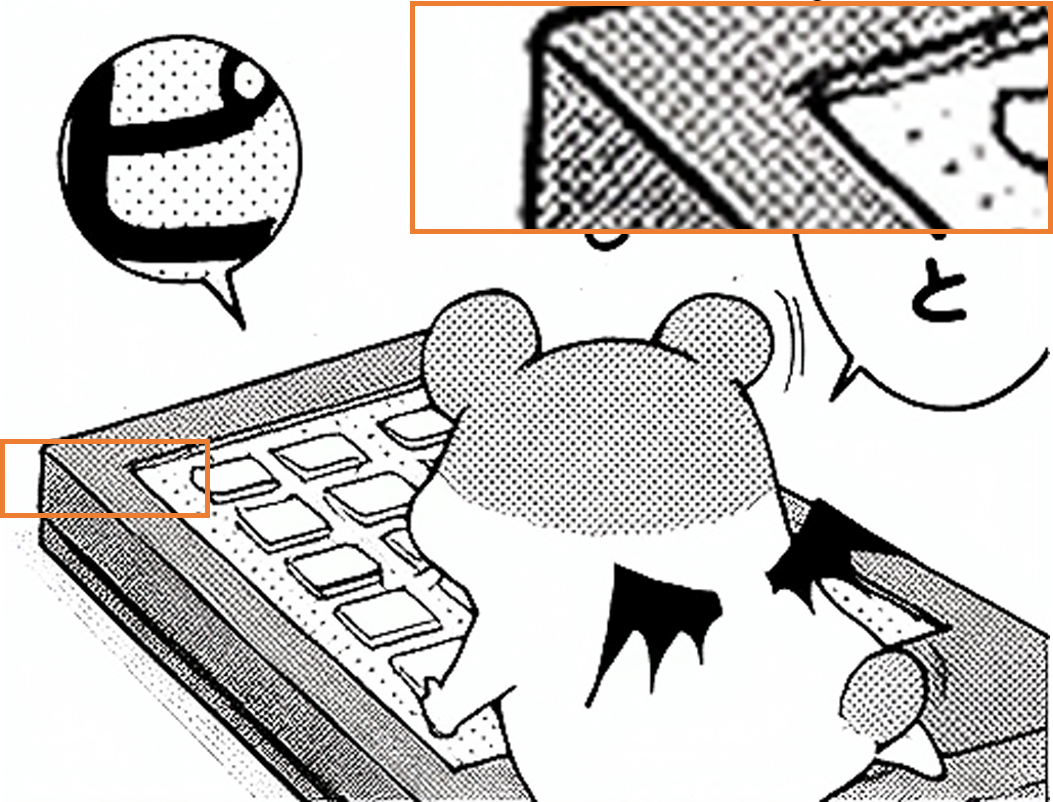}\hfil
        \includegraphics[width=0.48\linewidth]{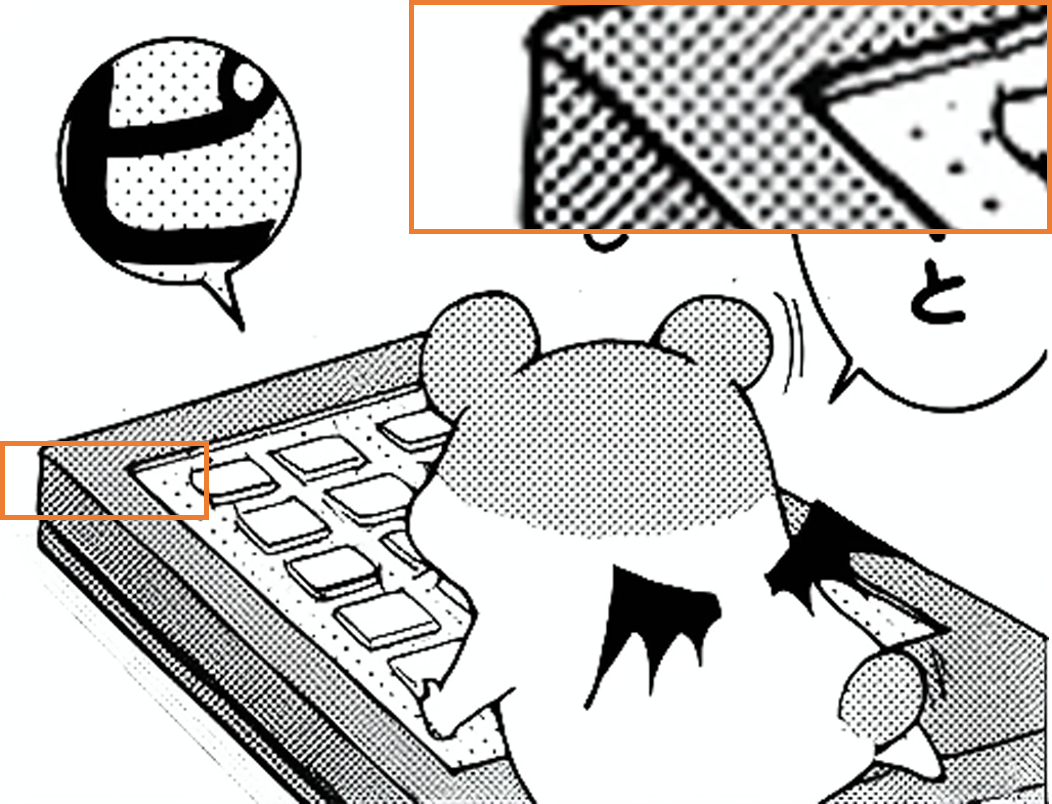}
    \end{minipage}
    \begin{tabularx}{0.481\textwidth}{*{2}{>{\centering\arraybackslash}X}}
            (c) Finetuned with LPIP loss &
            (d) Finetuned without LPIP loss
    \end{tabularx}
    \vspace{-0.15in}
    \caption{Ablation on VAE decoder. The original VAE decoder fails to generate visually pleasant screetones. Finetuning the VAE decoder without LPIP loss produces the best results.}
    \label{fig:vae}
\end{figure}

With the prepared training data, we finetune both denoise diffusion U-Net and the VAE decoder components in stable diffusion. The effectiveness of finetuning the latent diffusion is shown in Figure~\ref{fig:latentdiffusion}. The original latent diffusion model fails to generate high-frequency screentones due to the lack of training data, no matter the generation is conditioned on a sketch or an intensity map (Figure~\ref{fig:latentdiffusion}(b)\&(c)). Finetuning the latent diffusion enables the generation of high-frequency screentones, but conditioning on sketch still leads to bland unshaded screentones (Figure~\ref{fig:latentdiffusion}(d)). With intensity conditioning, our finetuned latent diffusion generates manga with shaded high-frequency screentones (Figure~\ref{fig:latentdiffusion}(e)). However, finetuning the latent diffusion alone still fails to generate visually pleasant screentones, as shown in Figure~\ref{fig:vae}(b). So we further finetune the decoder by following the paradigm of sd-vae-ft-mse\footnote{https://huggingface.co/stabilityai/sd-vae-ft-mse} that uses a weighted sum of adversarial loss~\cite{goodfellow2020generative}, MSE loss, and LPIP loss~\cite{zhang2018perceptual} as the objective function. Unfortunately, the result is still unsatisfactory after the finetuning (Figure~\ref{fig:vae}(c)). We find that LPIP loss degrades screentone reconstruction, probably due to the domain shift between manga and natural images. Therefore, we further propose to remove the LPIP loss from the objective function, which significantly enhances the quality of the generated screentones (Figure~\ref{fig:vae}(d)).

\subsection{Adaptive scaling}

\begin{figure}[!t]
    \begin{minipage}[b]{\linewidth}
        \includegraphics[width=0.247\linewidth]{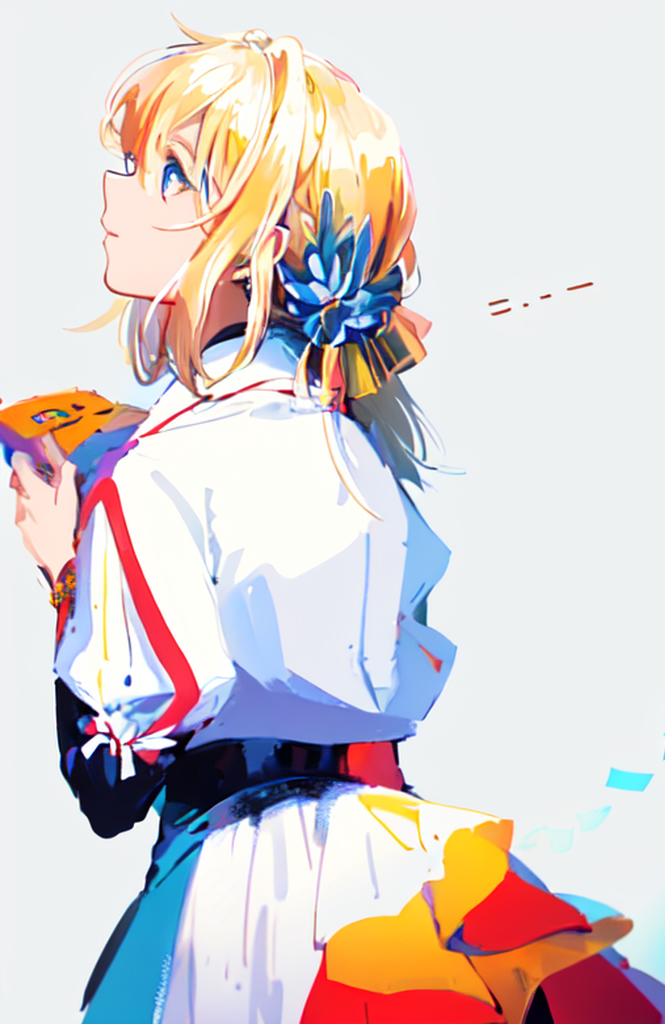}\hfil
        \includegraphics[width=0.247\linewidth]{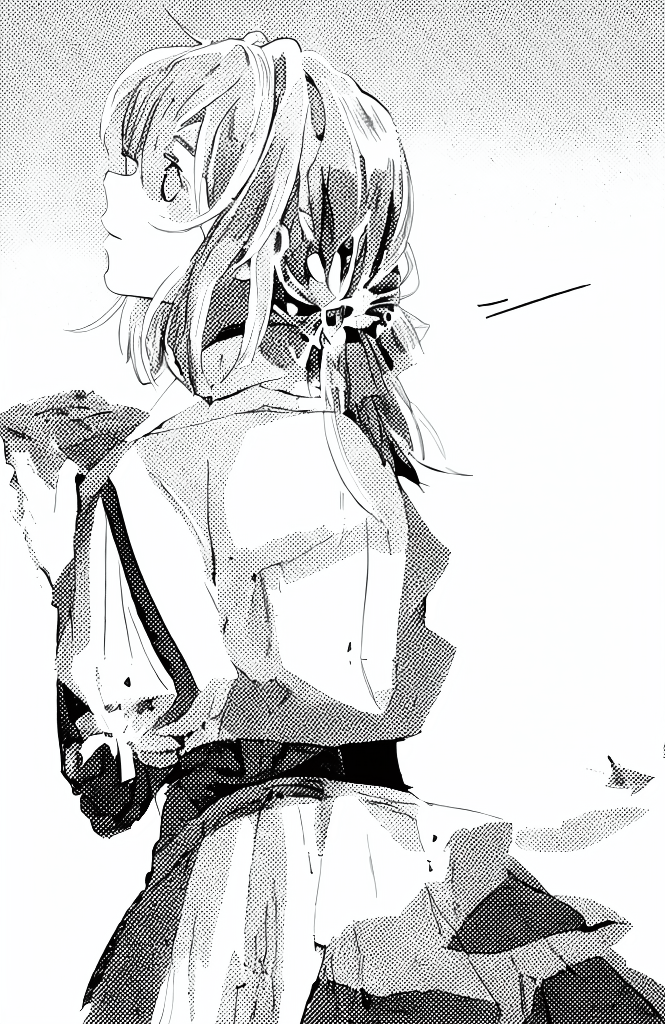}\hfil
        \includegraphics[width=0.247\linewidth]{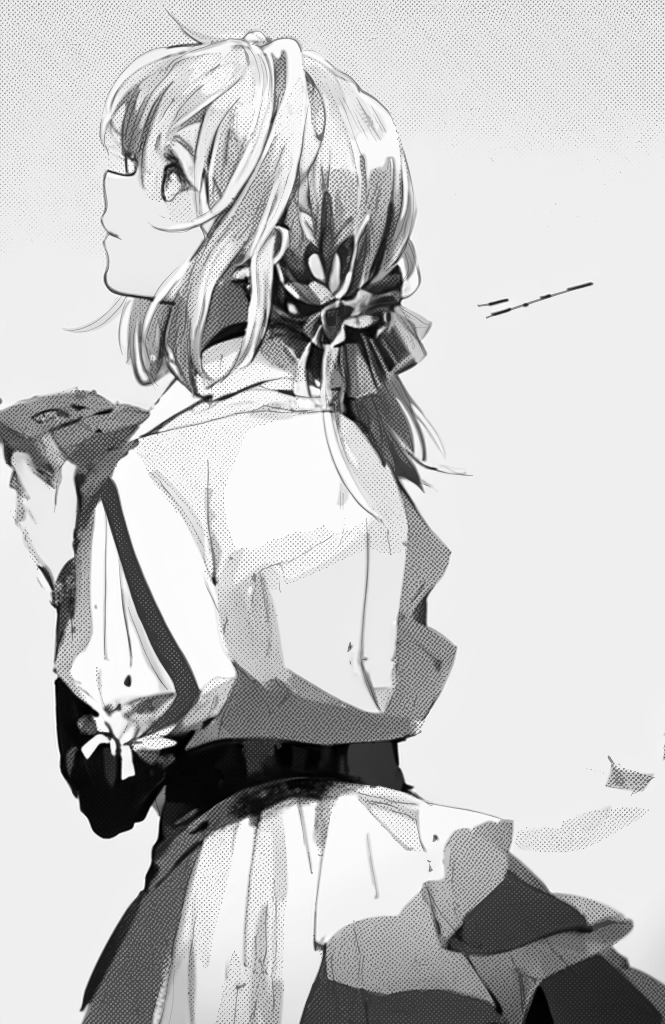}\hfil
        \includegraphics[width=0.247\linewidth]{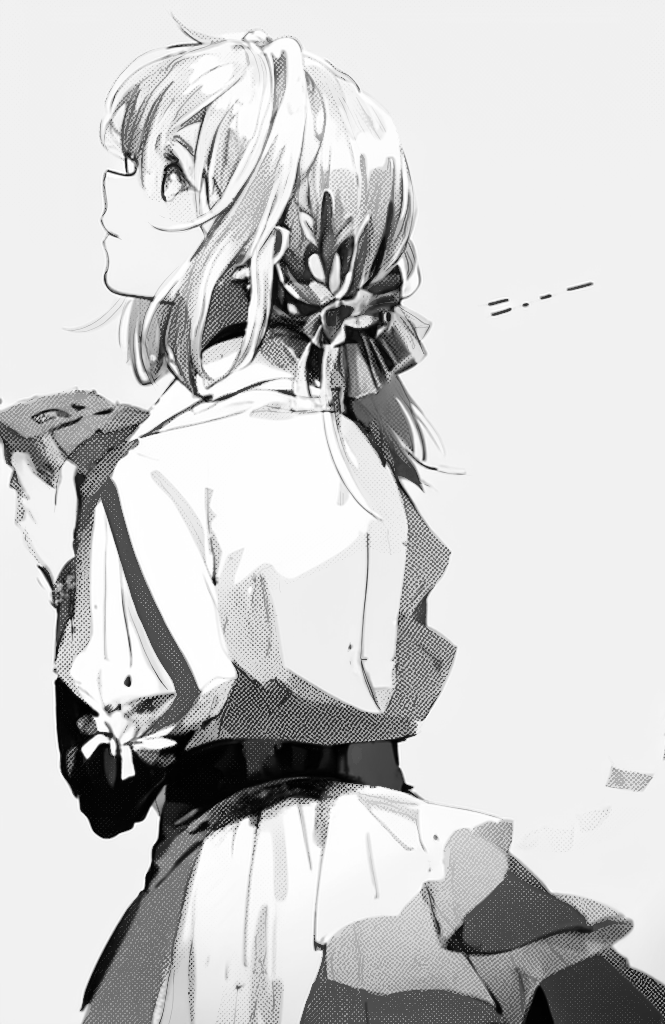}
    \end{minipage}
    \begin{tabularx}{\linewidth}{*{4}{>{\centering\arraybackslash}X}}
            (a) Color illustration &
            (b) Rough manga &
            (c) Scaling on L &
            (d) Scaling on S
    \end{tabularx}
    \vspace{-0.15in}
    \caption{Example of adaptive scaling on lightness (L) and saturation (S) respectively. Scaling on saturation achieves better results by preserving the shading of the color illustration.}
    \label{fig:compare_scaling}
\end{figure}

\begin{figure*}[!t]
    \begin{minipage}[b]{\linewidth}
        \includegraphics[width=0.245\linewidth, height=0.2\linewidth]{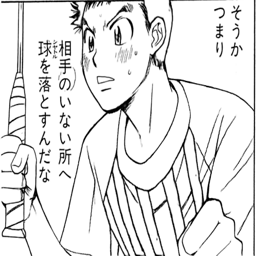}\hfil
        \includegraphics[width=0.245\linewidth, height=0.2\linewidth]{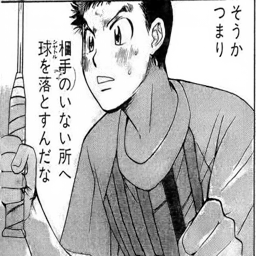}\hfil
        \includegraphics[width=0.245\linewidth, height=0.2\linewidth]{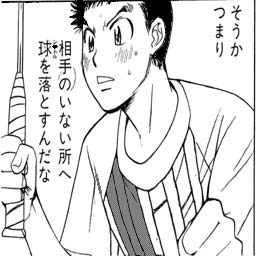}\hfil
        \includegraphics[width=0.245\linewidth, height=0.2\linewidth]{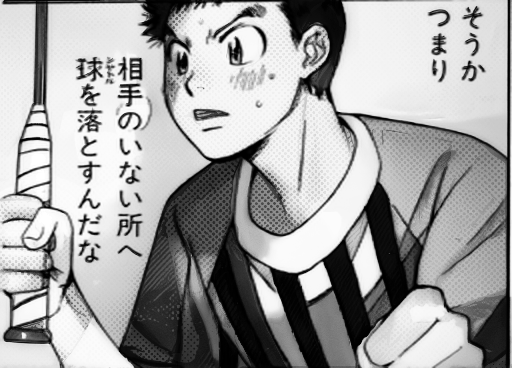}
    \end{minipage}
    \begin{minipage}[b]{\linewidth}
        \includegraphics[width=0.245\linewidth]{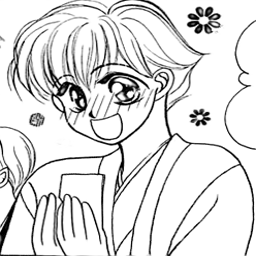}\hfil
        \includegraphics[width=0.245\linewidth]{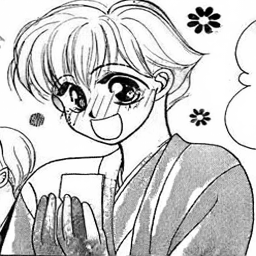}\hfil
        \includegraphics[width=0.245\linewidth]{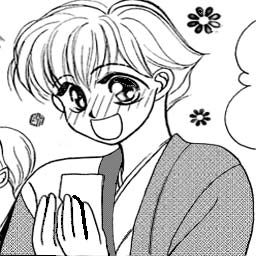}\hfil
        \includegraphics[width=0.245\linewidth, height=0.245\linewidth]{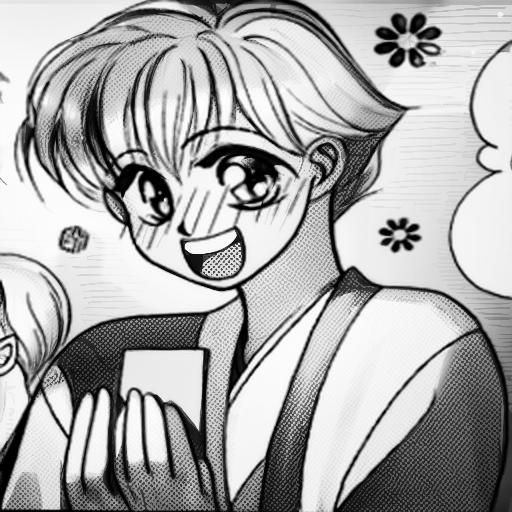}
    \end{minipage}
    \begin{tabularx}{\textwidth}{*{4}{>{\centering\arraybackslash}X}}
            Input & 
            pix2pix~\cite{pix2pix} & 
            Screentone Synthesis~\cite{tsubota2019synthesis} &
            Ours
    \end{tabularx}
    \caption{Comparisons with state-of-the-art sketch-to-manga methods.}
    \label{fig:sketch2manga}
\end{figure*}

\begin{table*}[!t]
    \centering
    \begin{tabular}{cc|cccc}
        \hline
         Sketch-to-manga & Preferred & Illustration-to-manga & Rank 1 & Rank 2 & Rank 3\\\hline
         Screentone Synthesis~\cite{tsubota2019synthesis} & 16.2\% & ScreenVAE~\cite{screenvae} & 3.0\% & 6.4\% & 90.6\%\\
         Ours & \textbf{83.8\%} & Mimicking Manga~\cite{Manga2021zhang} & 32.9\% & 66.5\% & 0.6\%\\
         & & Ours & \textbf{64.1\%} & 27.1\% & 8.8\%\\
         \hline
    \end{tabular}
    \caption{User studies on sketch-to-manga and color-to-manga.}
    \label{tab:user_study}
\end{table*}

Our finetuned intensity-conditioned diffusion model produces manga with shaded high-frequency screentones, but also with subtle structure distortion and unnatural shading introduced by the diffusion process (as shown in the area around the eyes in Figure~\ref{fig:compare_scaling}(b)), we propose a color-screentone integration scheme to integrate the generated color illustrations and the generated rough manga with shaded high-frequency screentones to produce the final manga image.

Given an input sketch $I_s$, we denote the generated color illustration as $I_c$ and its grayscale version as $I_g$. The rough manga image with shaded high-frequency screentones generated by the diffusion model is denoted as $I_r$. We first apply K-Means clustering to the generated color illustration $I_c$ to obtain a set of regions where pixels in each region are connected and with similar colors. For each identified region $R$, the distribution of screentones in the generated rough manga image is utilized to decide how likely this region should be screentoned. We propose that if the standard deviation of the region in the generated rough manga image $\sigma^R$ is large, it is very likely that this region $R$ should be screentoned, and vice versa. Based on this assumption, we calculate a scaling range $[s^R_{low},s^R_{high}]$ defined by two scaling factors $s^R_{low}$ and $s^R_{high}$ for each region $R$ based on its standard deviation $\sigma^R$ as
$$
\begin{aligned}
s^R_{low} &=& 1 &- w_{low} \cdot \sigma^R \\
s^R_{high} &=& 1 &+ w_{high} \cdot \sigma^R 
\end{aligned}
$$
Here, the scaling factors $s^R_{low}$ and $s^R_{high}$ indicate the lowest bound and the highest bound of the scaling range respectively. $w_{low}$ and $w_{high}$ are two weighting parameters and empirically set to 0.08 and 0.16 in all our experiments.
With this pair of region-based scaling factors $s^R_{low}$ and $s^R_{high}$, we are able to scale the colors of all pixels inside the region $R$ in the color illustration $I_c$ in an adaptive way.

We propose two adaptive scaling strategies by scaling the colors on the lightness channel $L_c$ and saturation channel $S_c$ respectively. In the following, adaptive scaling on the saturation channel is described, while adaptive scaling on lightness channel can be obtained in a similar manner. For each region $R$ of $I_r$, we denote the minimum and maximum intensity values of this region as $\min(I^R_r)$ and $\max(I^R_r)$ respectively. The value of a pixel $p$ in the saturation channel $S_c$ is updated by multiplying a scaling factor $s$, i.e., $S_c[p] = S_c[p] \cdot s$, where $s$ is calculated as
$$
s = s^R_{high} - \frac{I_r[p]-\min(I^R_r)}{\max(I^R_r)-\min(I^R_r)}\cdot\left(s^R_{high} - s^R_{low}\right)
$$

Figure~\ref{fig:compare_scaling} shows an example where both results of scaling on lightness channel and scaling on saturation channel are presented. We can observe that scaling on lightness channel can better preserve the generated screentones, while scaling on saturation can better preserve the shading in the color illustration. Empirically, scaling on saturation provides more natural shading, similar to that in real-world manga. Therefore, all results we provide in the paper are acquired by region-based adaptive scaling on saturation.

\begin{figure*}[!t]
    \centering
    \begin{minipage}[b]{\linewidth}
        \includegraphics[width=0.248\linewidth]{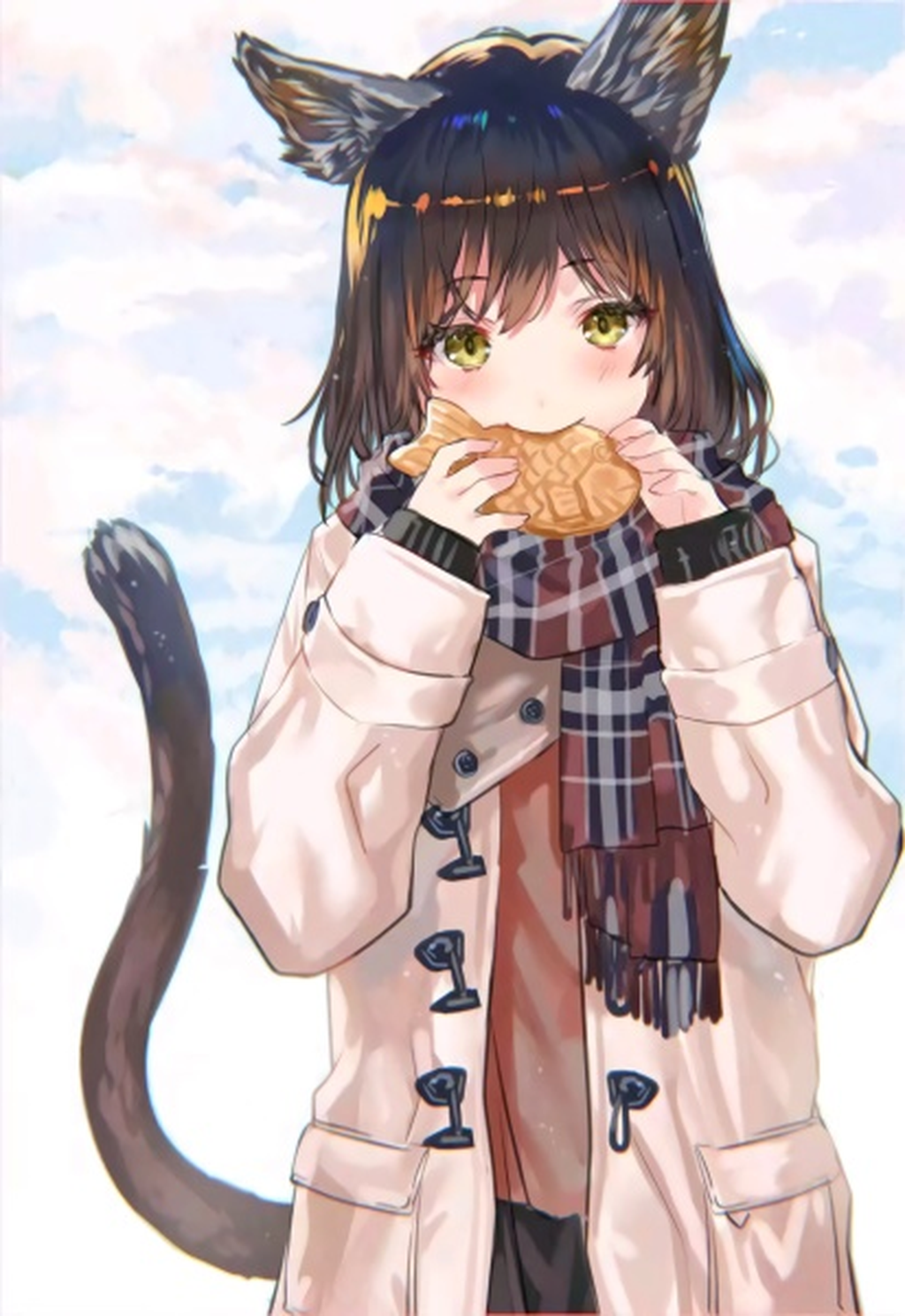}\hfil
        \includegraphics[width=0.248\linewidth]{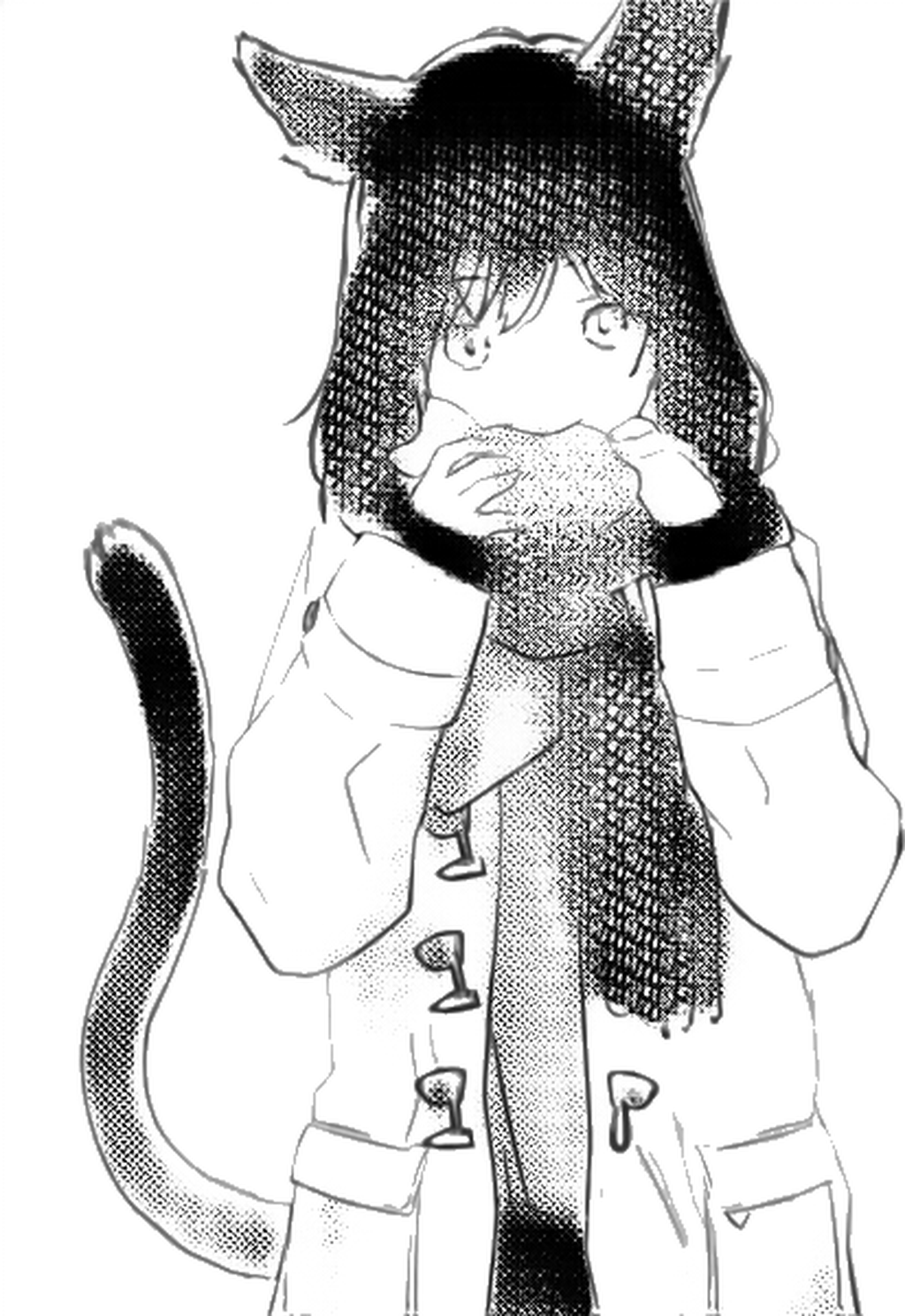}\hfil
        \includegraphics[width=0.248\linewidth]{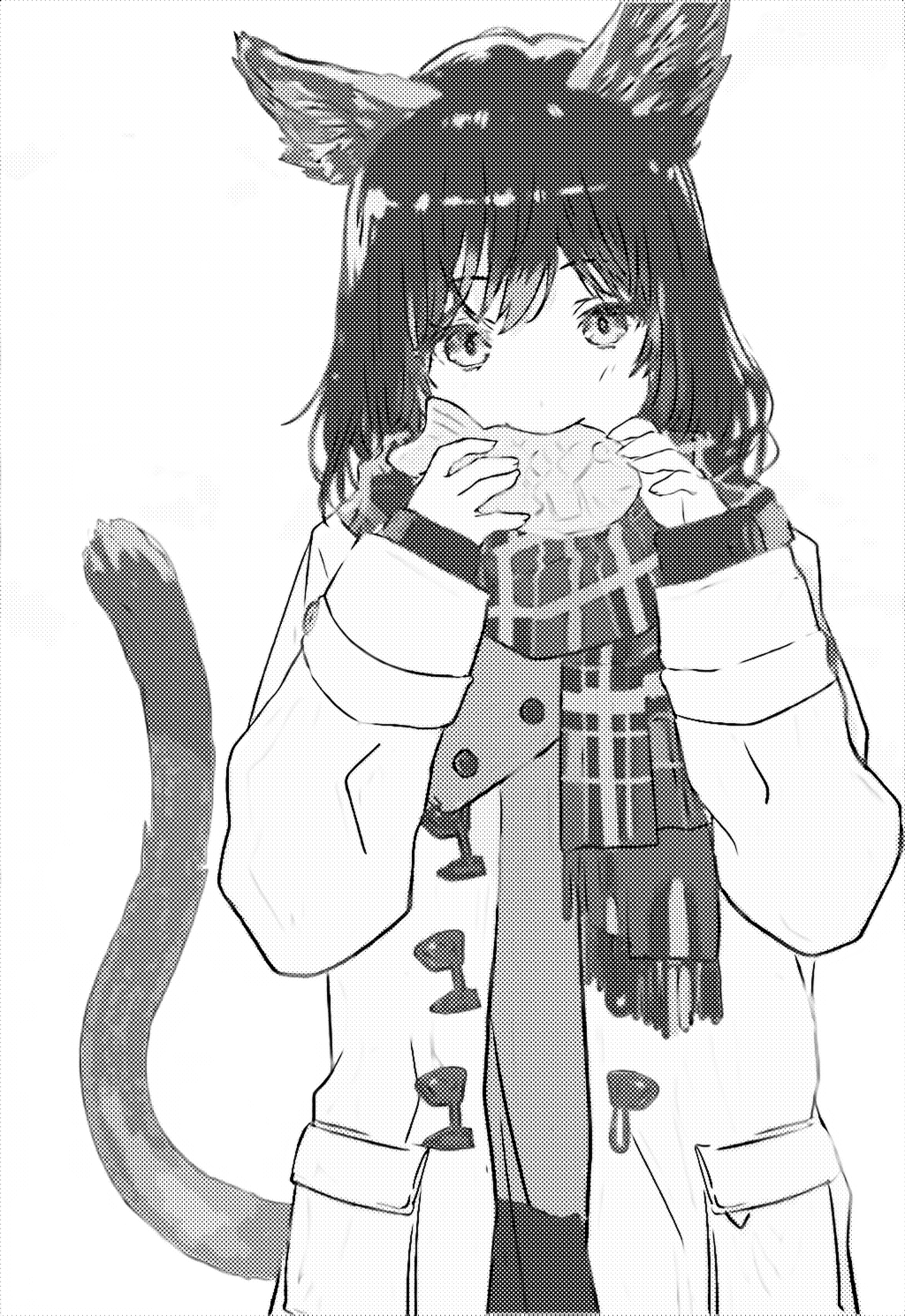}\hfil
        \includegraphics[width=0.248\linewidth]{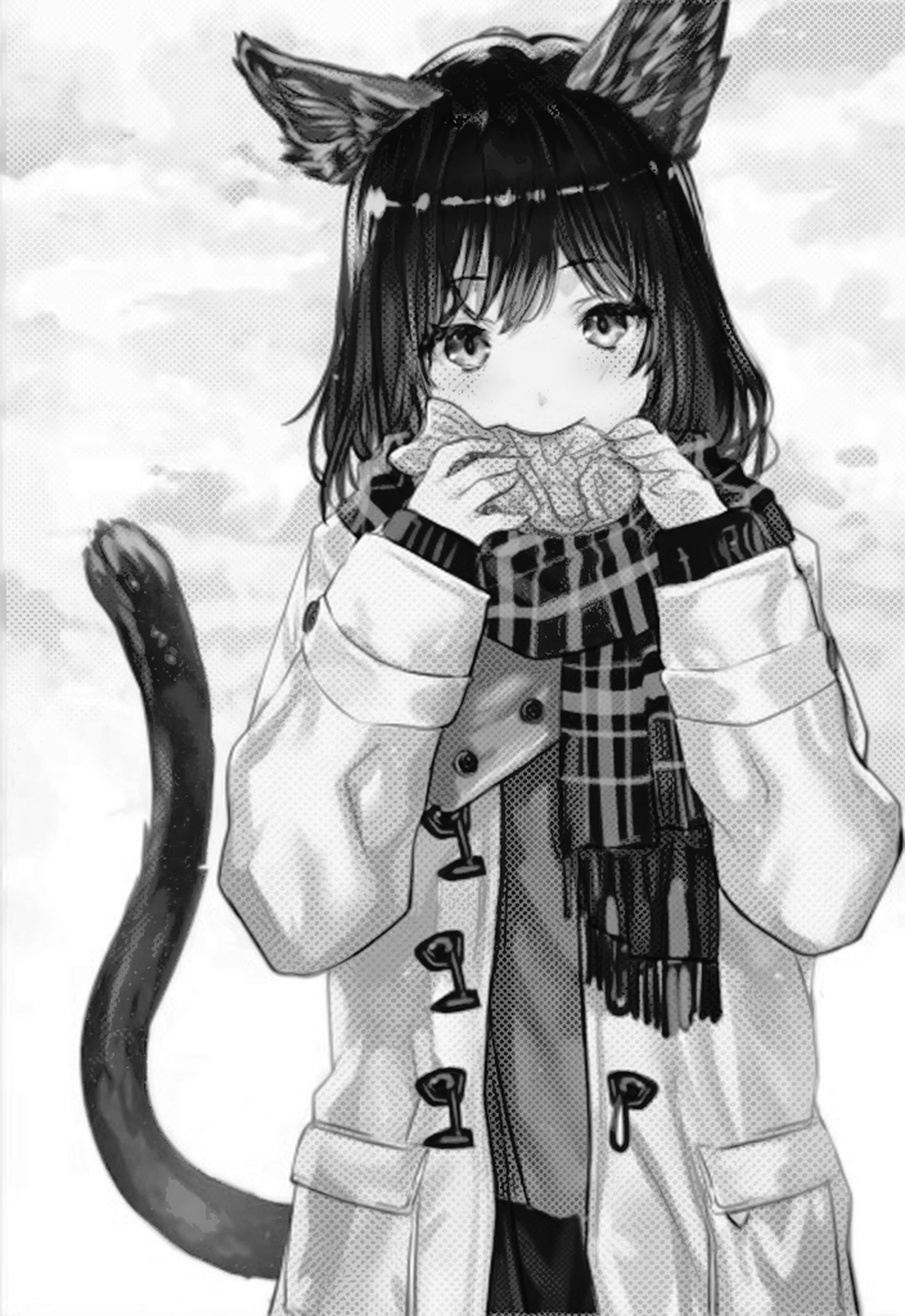}
    \end{minipage}
    \begin{minipage}[b]{\linewidth}
        \includegraphics[width=0.248\linewidth]{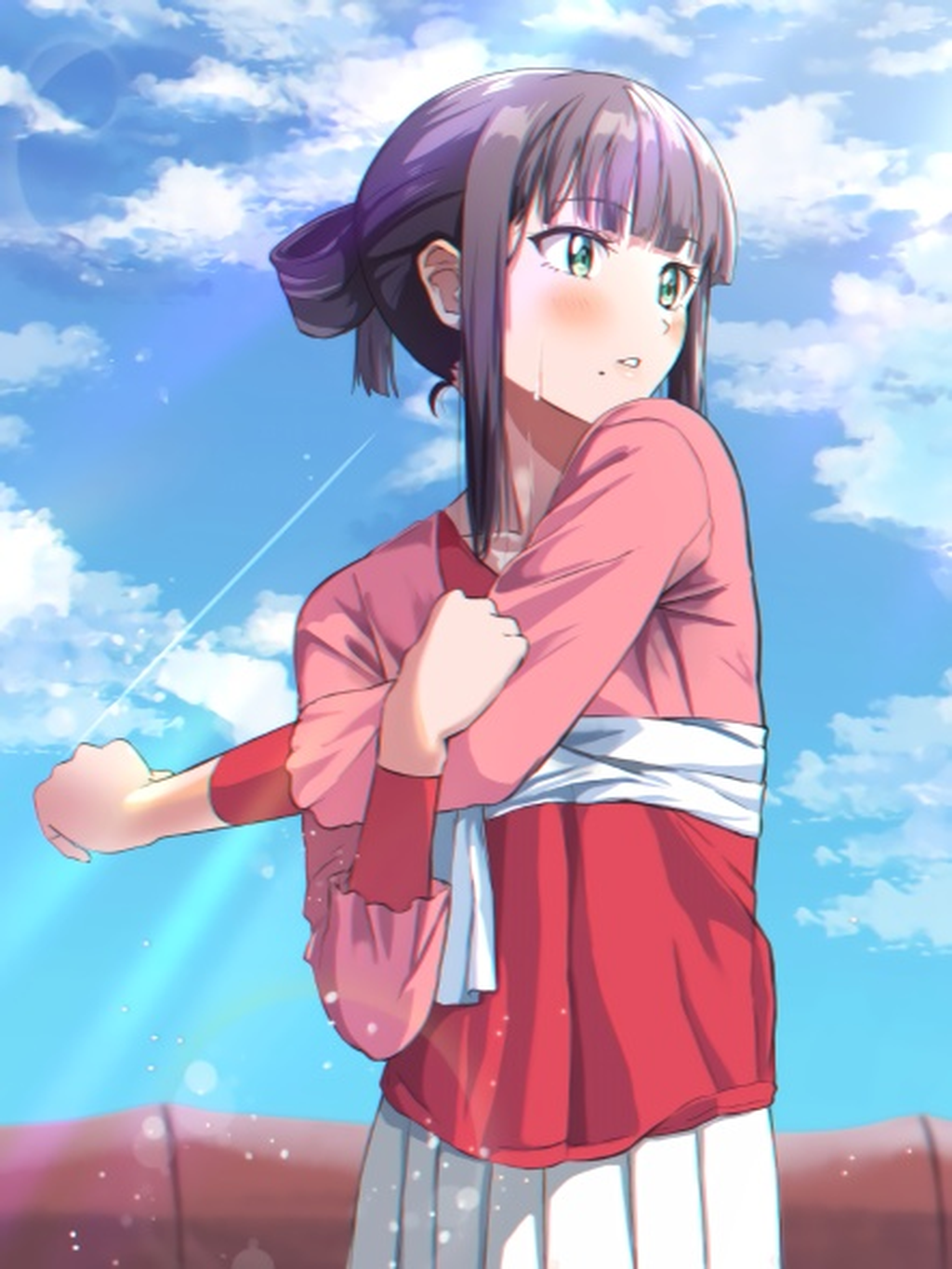}\hfil
        \includegraphics[width=0.248\linewidth]{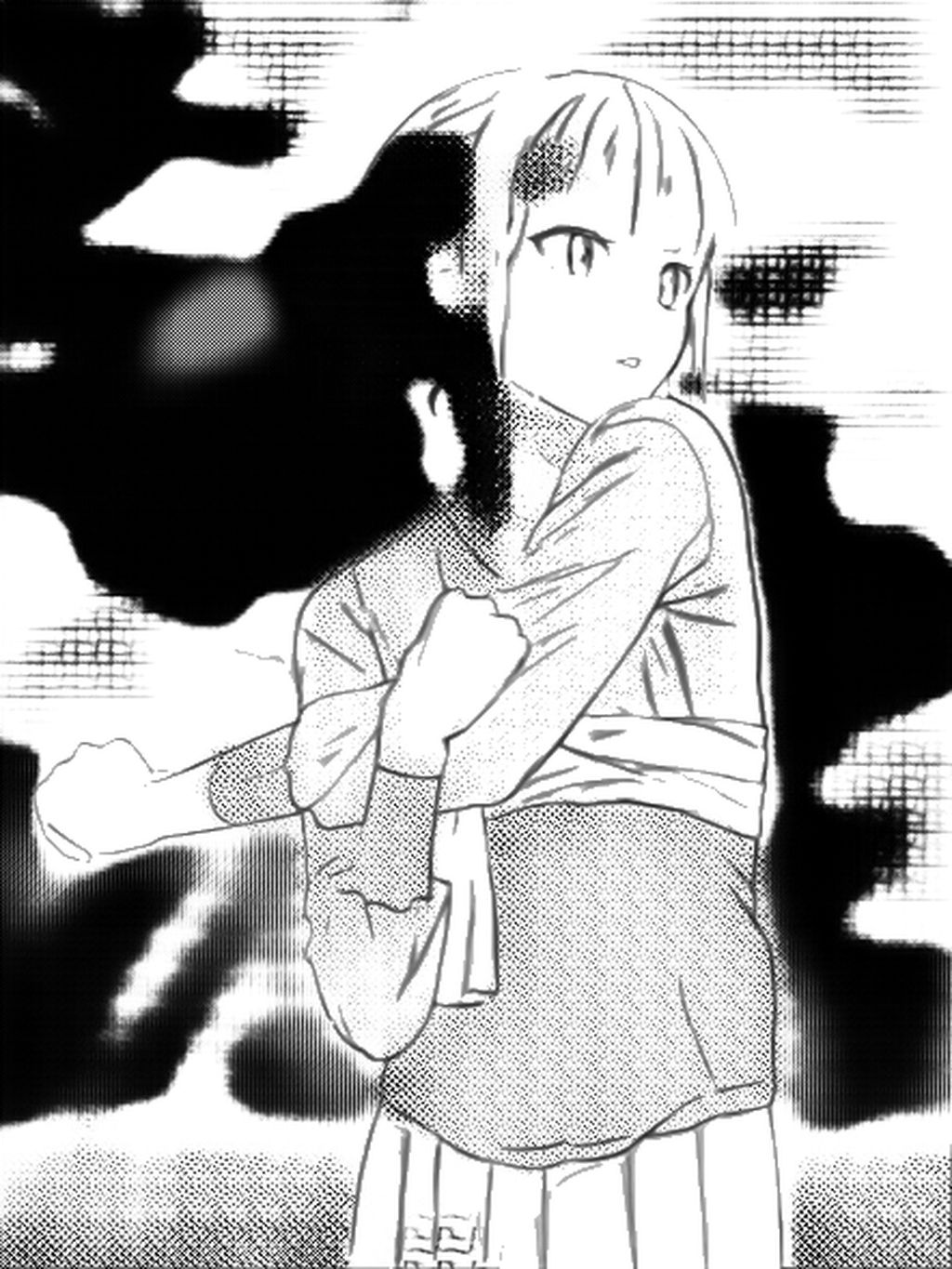}\hfil
        \includegraphics[width=0.248\linewidth]{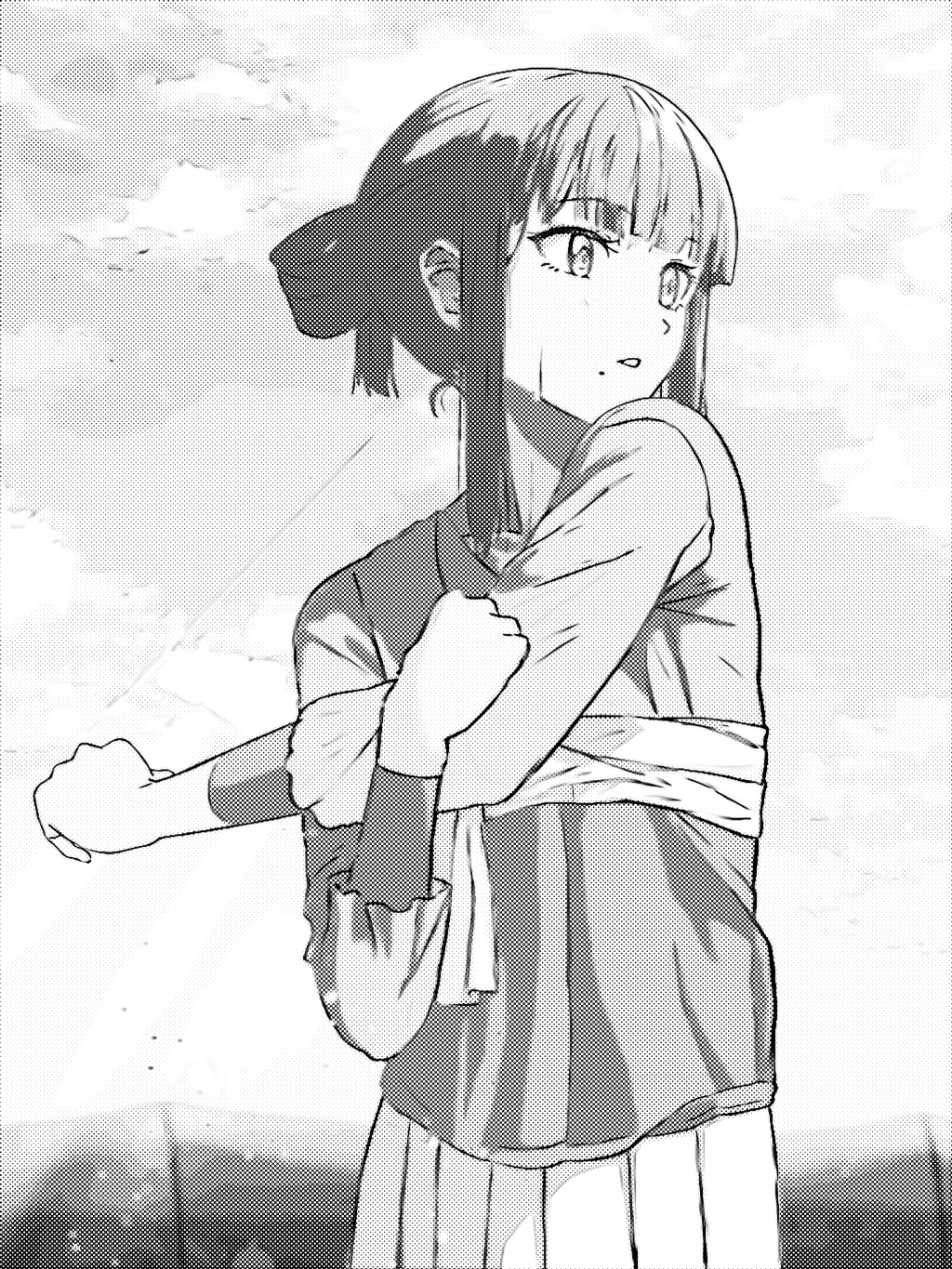}\hfil
        \includegraphics[width=0.248\linewidth]{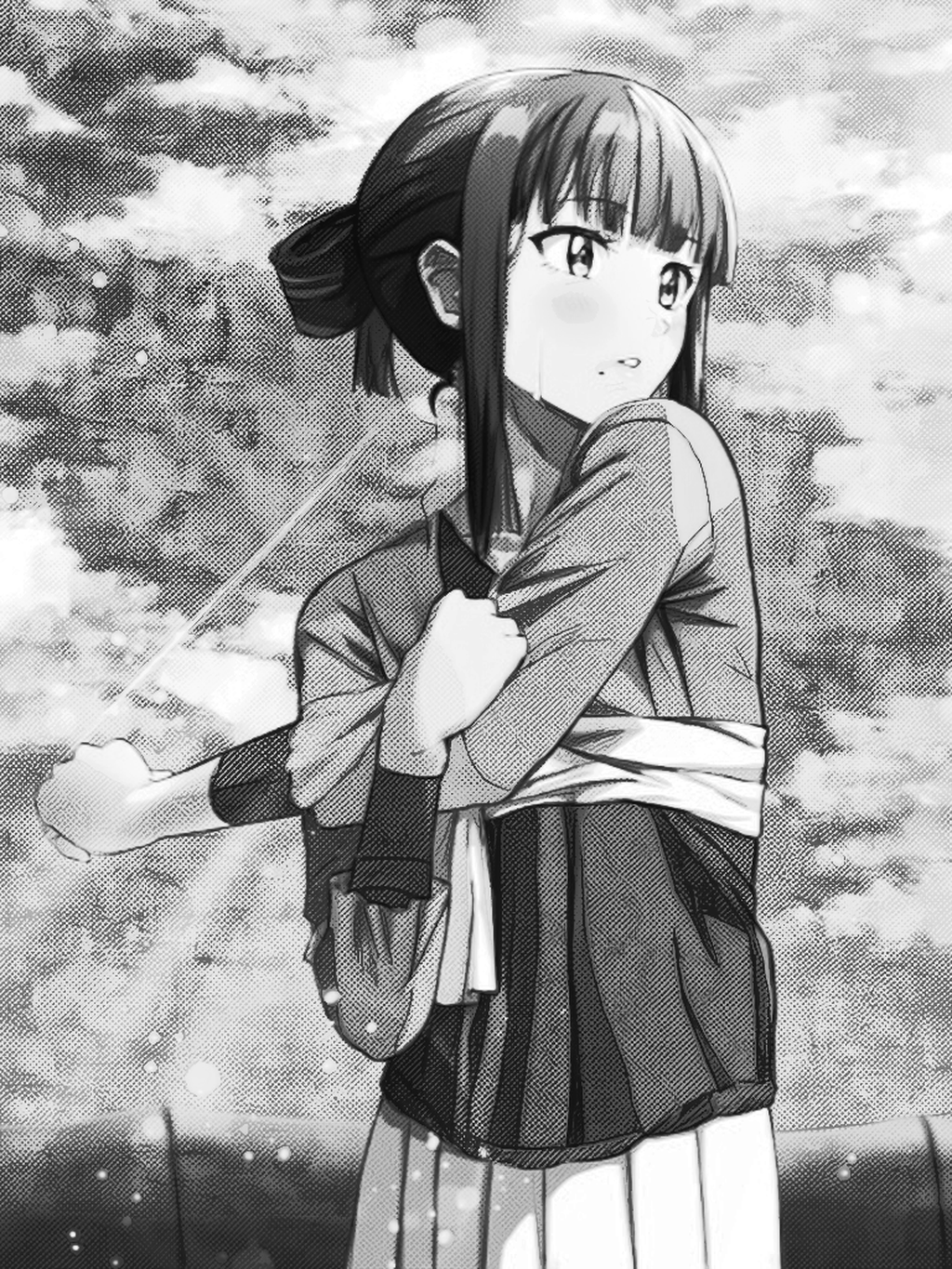}
    \end{minipage}
    \begin{minipage}[b]{\linewidth}
        \includegraphics[width=0.248\linewidth]{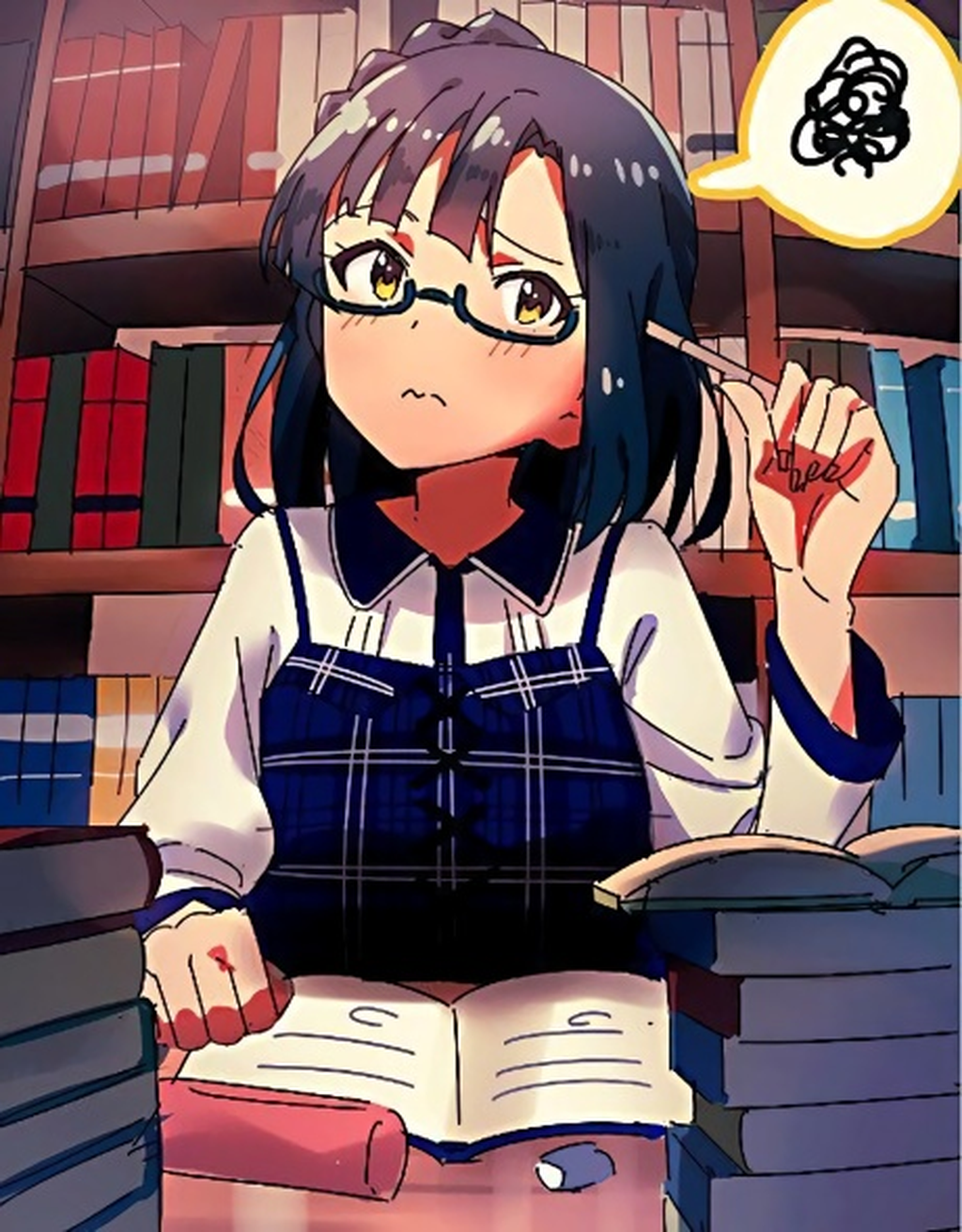}\hfil
        \includegraphics[width=0.248\linewidth]{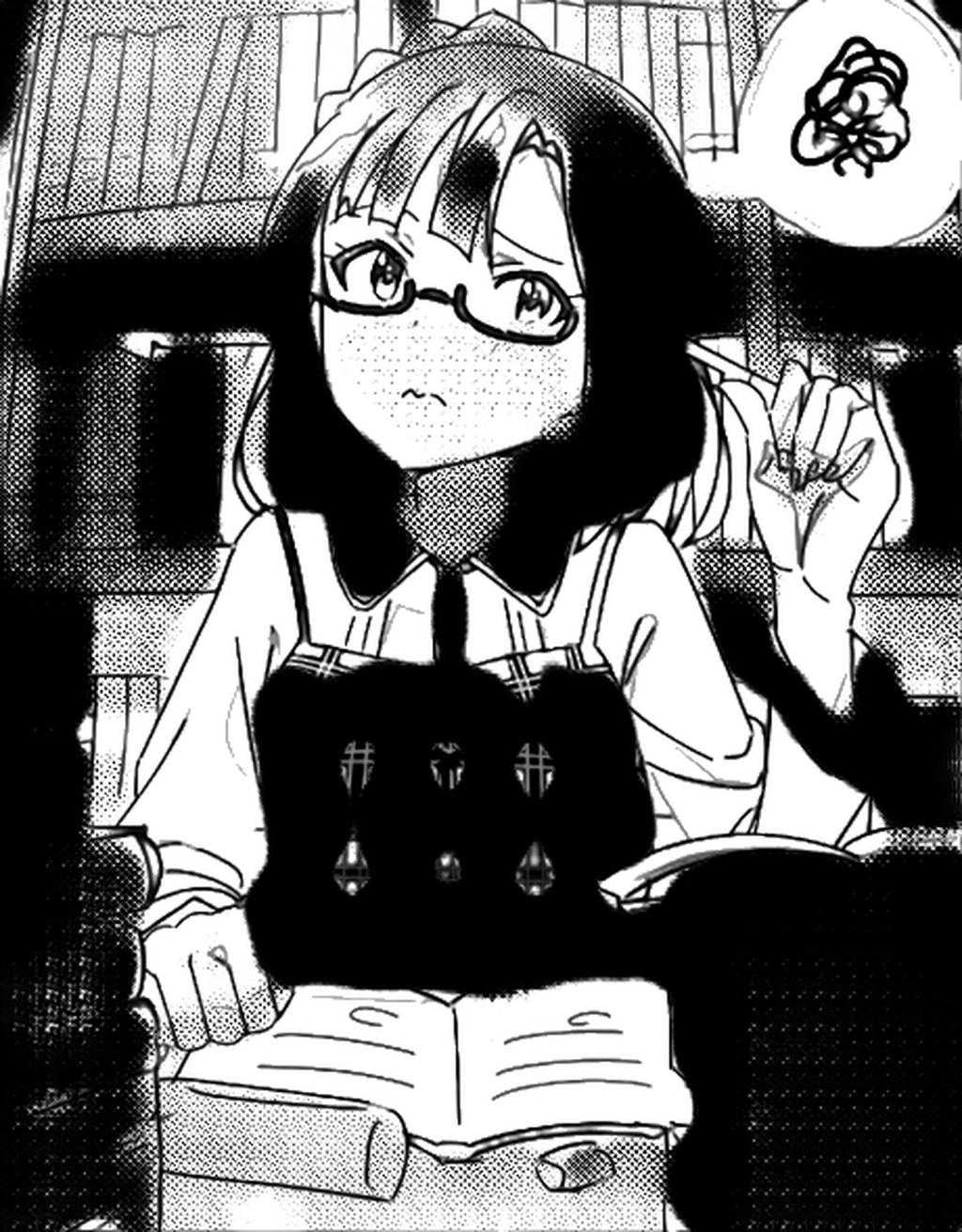}\hfil
        \includegraphics[width=0.248\linewidth]{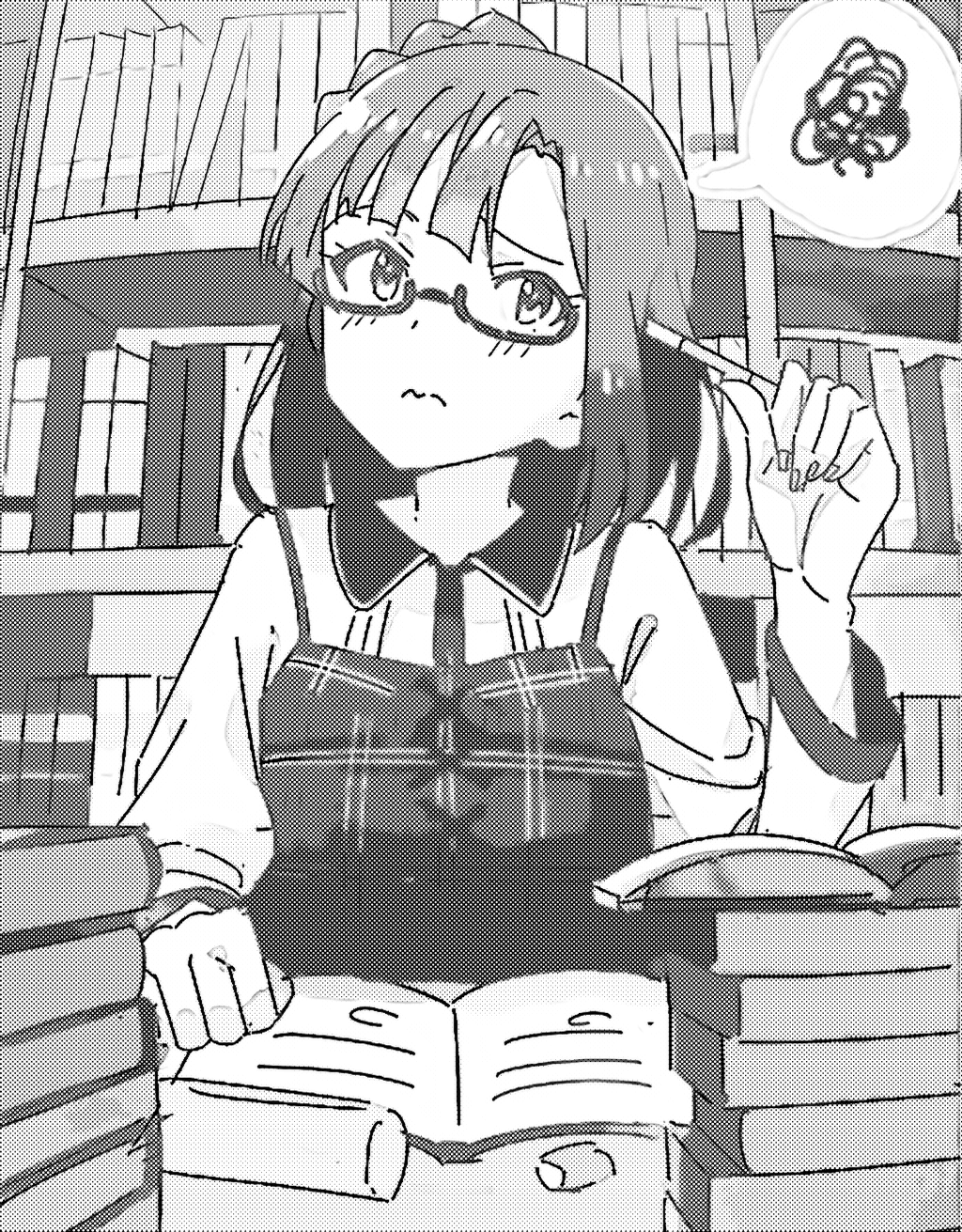}\hfil
        \includegraphics[width=0.248\linewidth]{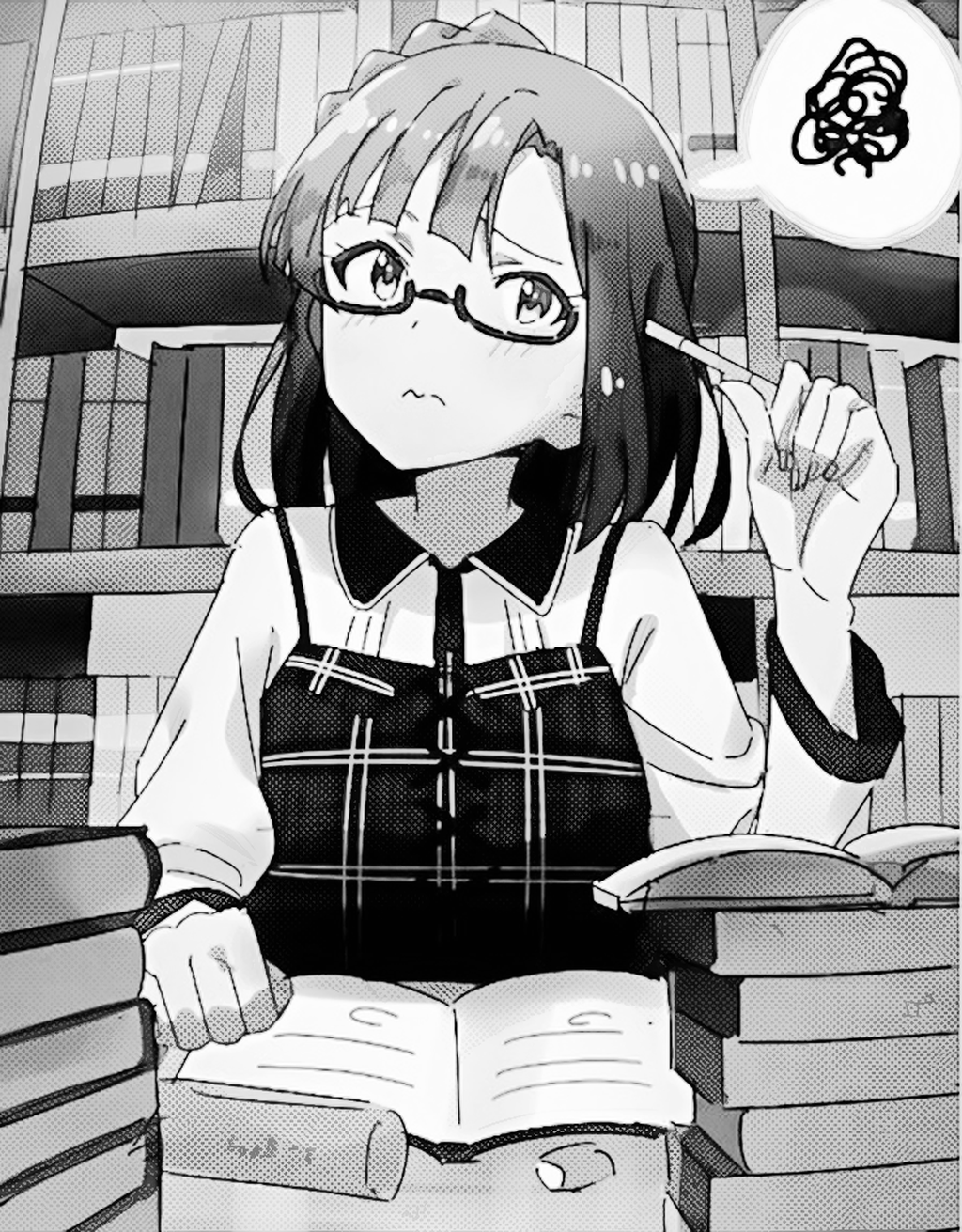}
    \end{minipage}
    \begin{tabularx}{\linewidth}{*{4}{>{\centering\arraybackslash}X}}
            (a) Input & 
            (c) ScreenVAE~\cite{screenvae} &
            (b) Mimicking Manga~\cite{Manga2021zhang} & 
            (d) Ours
    \end{tabularx}
    \caption{Comparisons with state-of-the-art illustration-to-manga methods.}
    \label{fig:illustration2manga}
\end{figure*}

\section{Results}
To validate the effectiveness of our method, we compare with the state-of-the-art sketch-to-manga methods and illustration-to-manga methods. For sketch-to-manga, there is only one automatic method proposed so far, namely Screentone Synthesis~\cite{tsubota2019synthesis}. We also compare with the image-to-image translation model pix2pix~\cite{pix2pix}. For illustration-to-manga, we compare with two state-of-the-art methods, including ScreenVAE~\cite{screenvae} and Mimicking Manga~\cite{Manga2021zhang}.

\vspace{0.1in}
\noindent\textbf{Visual Comparisons} Figure~\ref{fig:sketch2manga} shows the comparisons with existing sketch-to-manga methods. Pix2pix~\cite{pix2pix} generates blurred unshaded screentones with unnatural placement. Screentone Synthesis~\cite{tsubota2019synthesis} generate relatively high-frequency screentones, but are still unshaded with unnatural placement. In comparisons, our method generates high-quality manga images with shaded high-frequency screentones.

Figure~\ref{fig:illustration2manga} shows the comparisons with existing color-to-manga methods. ScreenVAE~\cite{screenvae} generates high-frequency screentones with proper shading transitions in flat or almost flat color regions. However, this model faces challenge in regions with rich colors, and leading to frequently occurred black regions in the generated manga image. Mimicking manga~\cite{Manga2021zhang} generates high-quality screentones since they rely on screentone exemplars to synthesize the screentones in the output image. However, the screentones are unnaturally shaded, leading to consistent low-contrast manga output. In contrast, our method achieves significantly better results than existing methods even for color illustration inputs.

\vspace{0.1in}
\noindent\textbf{User Studies} We conduct user studies with 20 unseen sketches for sketch-to-manga and 20 unseen illustrations for illustration-to-manga from Danbooru~\cite{danbooru2021}. 17 participants are invited. For sketch-to-manga, since there is only one competitor, the users are asked to choose their preferred one from the two results. For illustration-to-manga, since there are two competitors, the users are asked to rank the three results based on their preferences. Table~\ref{tab:user_study} shows the results where our method significantly outperforms the rest.

\section{Conclusion}

In this paper, we proposed a novel sketch to manga framework that utilized color illustration as the intermediary. A tailored manga generation diffusion model conditioned on intensity is proposed by finetuning on high-resolution manga data with a tailored loss function. An adaptive scaling method is further proposed to integrate the generated rough manga with high-quality screentones into the color illustration.

The proposed method could be further extended to add screentones to a full manga page or a line cartoon. We would further explore such application while maintaining the consistency among different panels in a manga page or among consecutive frames in a line cartoon video.

\bibliographystyle{IEEEbib}
\bibliography{template}

\end{document}